\pdfoutput=1
\documentclass[sigconf]{acmart}

\usepackage{latexsym}
\usepackage{algorithmic}
\usepackage{amsmath}
\usepackage{esvect}
\usepackage{latexsym}
\usepackage{boxedminipage} 
\usepackage{CJK}
\usepackage{enumitem}
\usepackage{mathrsfs}
\usepackage{amsfonts}
\usepackage{bm}
\usepackage{colortbl}
\usepackage{graphicx}
\usepackage{float}
\usepackage{subfigure}
\usepackage{soul}
\usepackage{booktabs} 
\usepackage{url}
\usepackage{amsmath}
\usepackage{xcolor}
\usepackage{algorithm}
\usepackage{makecell}
\usepackage{xspace}
\usepackage{array}
\usepackage[normalem]{ulem}
\usepackage{footmisc}
\usepackage{colortbl}
\usepackage{epstopdf}
\usepackage{setspace}
\usepackage{pifont}
\usepackage{rotating,tabularx}
\usepackage{multirow}

\definecolor{gred}{RGB}{219,68,55}
\definecolor{gblue}{RGB}{66,133,244}
\definecolor{gyellow}{RGB}{244,180,0}
\definecolor{ggreen}{RGB}{15,157,88}
\definecolor{ggrey}{RGB}{115,115,115}

\newcommand{\error}[1]{\textcolor{gred}{\textbf{#1}}} 
\newcommand{\fph}[1]{\textcolor{gblue}{\textbf{#1}}} 
\newcommand{\reph}[1]{\textcolor{ggreen}{\textbf{#1}}} 

\AtBeginDocument{%
  \providecommand\BibTeX{{%
    \normalfont B\kern-0.5em{\scshape i\kern-0.25em b}\kern-0.8em\TeX}}}

\copyrightyear{2023}
\acmYear{2023}
\setcopyright{rightsretained}
\acmConference[SIGIR '23]{Proceedings of the 46th International ACM SIGIR Conference on Research and Development in Information Retrieval}{July 23--27, 2023}{Taipei, Taiwan}
\acmBooktitle{Proceedings of the 46th International ACM SIGIR Conference on Research and Development in Information Retrieval (SIGIR '23), July 23--27, 2023, Taipei, Taiwan}\acmDOI{10.1145/3539618.3591630}
\acmISBN{978-1-4503-9408-6/23/07}
\makeatletter
\patchcmd{\maketitle}{\@copyrightpermission}{
   \begin{minipage}{0.3\columnwidth}
     \href{http://creativecommons.org/licenses/by/4.0/}{\includegraphics[width=0.70\textwidth]{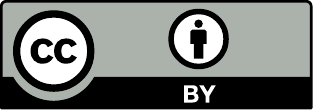}}
   \end{minipage}\hfill
   \begin{minipage}{0.7\columnwidth}
     \href{http://creativecommons.org/licenses/by/4.0/}{This work is licensed under a Creative Commons Attribution International 4.0 License.}
   \end{minipage}
  
}{}{}

\makeatother
\begin{document}

\title{A Topic-aware Summarization Framework with Different Modal Side Information}

\author{Xiuying Chen}
\authornotemark[1]
\affiliation{%
  \institution{CBRC, KAUST\\
  CEMSE, KAUST}
  \country{Saudi Arabia}
}
\email{xiuying.chen@kaust.edu.sa}

\author{Mingzhe Li}
\authornotemark[1]
\affiliation{%
  \institution{Ant Group}
  \country{China}
}
\email{limingzhe.lmz@antgroup.com}

\author{Shen Gao}
\affiliation{%
  \institution{ Shandong University}
  \country{China}
}
\email{shengao@sdu.edu.cn}

\author{Xin Cheng}
\affiliation{%
  \institution{ Peking University}
  \country{China}
}
\email{chengxin1998@stu.pku.edu.cn}

\author{Qiang Yang}
\affiliation{%
  \institution{CBRC, KAUST\\
  CEMSE, KAUST}
  \country{Saudi Arabia}
}
\email{qiang.yang@kaust.edu.sa}

\author{Qishen Zhang}
\affiliation{%
  \institution{Ant Group}
  \country{China}
}
\email{qishen.zqs@antgroup.com}

\author{ Xin Gao$^{\dagger}$}
\affiliation{%
  \institution{CBRC, KAUST\\
  CEMSE, KAUST}
  \country{Saudi Arabia}
}
\email{xin.gao@kaust.edu.sa}

\author{Xiangliang Zhang$^{\dagger}$}
\affiliation{%
  \institution{University of Notre Dame\\CEMSE, KAUST
  }
  \country{USA}
}
\email{xzhang33@nd.edu}

\thanks{* Equal contribution.} 
\thanks{$\dagger$ Corresponding authors.} 

\def\authors{Xiuying Chen, Mingzhe Li, Shen Gao, Xin Cheng, Qiang Yang, Qishen Zhang, Xin Gao, Xiangliang Zhang}

\renewcommand{\shortauthors}{Xiuying Chen et al.}
\begin{abstract}
Automatic summarization plays an important role in the exponential document growth on the Web. 
On content websites such as CNN.com and WikiHow.com, there often exist various kinds of side information along with the main document for attention attraction and easier understanding, such as videos, images, and queries.
Such information can be used for better summarization, as they often explicitly or implicitly mention the essence of the article.
However, most of the existing side-aware summarization methods are designed to incorporate either single-modal or multi-modal side information, and cannot effectively adapt to each other. 
In this paper, we propose a general summarization framework, which can flexibly incorporate various modalities of side information.
The main challenges in designing a flexible summarization model with side information include: (1) the side information can be in textual or visual \textit{format}, and the model needs to align and unify it with the document into the same semantic space, (2) the side inputs can contain information from various \textit{aspects}, and the model should recognize the aspects useful for summarization.
To address these two challenges, we first propose a unified topic encoder, which jointly discovers latent topics from the document and various kinds of side information.
The learned topics flexibly bridge and guide the information flow between multiple inputs in a graph encoder through a topic-aware interaction. 
We secondly propose a triplet contrastive learning mechanism to align the single-modal or multi-modal information into a unified semantic space, where the \textit{summary} quality is enhanced by better understanding the \textit{document} and \textit{side information}.
Results show that our model significantly surpasses strong baselines  on three public single-modal or multi-modal benchmark summarization datasets.
\end{abstract}




\maketitle

	\section{Introduction}

\begin{figure*}[t]
		\centering
		\includegraphics[width=0.86\linewidth]{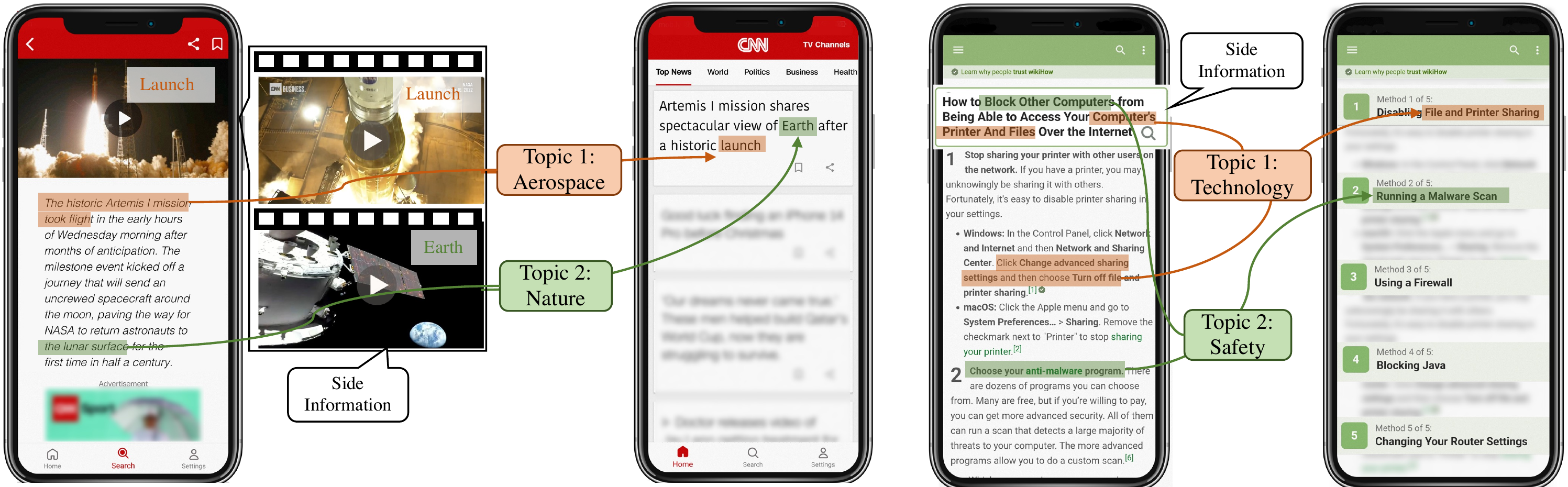}
		\caption{Articles with various side information and summary collected from the  CNN and WikiHow APPs.
		The side information (video and user query) can enhance the summarization performance.}
		\label{fig:intro}
	\end{figure*}

	The rapid growth of the World Wide Web has led to the flood of information across the Internet \cite{huang2019mist,shin2019sweg,ye2020leveraging}. 
    On content websites such as CNN.com, Twitter.com, and WikiHow.com, there are often corresponding images, videos, and side text along with the main document, which can attract readers' attention and help them understand the content better \cite{wang2018reconstruction,chen2018less,yan2019stat,qian2021generating}.
    Herein, we regard the auxiliary images/videos/text as \textbf{side information}.
    Since the side information frequently make reference to the article's main content explicitly or implicitly, such information can also be used to improve summarization quality, as shown by the  two examples from CNN and WikiHow Apps in Figure~\ref{fig:intro}.
    There is also other side information in real-world applications such as citation papers, summary templates, and reader comments, which are helpful for summarization~\cite{gao2021standard}.
    It is thus desired to extend text-based summarization models for taking advantage of the summarization clues  included in such side information.
	
	 There are previous works exploring utilizing side information from a specific domain.
	 For example, \citet{narayan2017neural} first proposed to utilize image captions to enhance summarization performance.
  Other textual side information such as citation papers~\cite{an2021enhancing}, reader comments~\cite{Gao2019Abstractive}, user queries~\cite{Koupaee2018WikiHowAL}, prototype templates~\cite{gao2019write} are also utilized in summarization tasks.
	 Recently, the benefits of visual information on summarization have also been explored.
	 To name a few, \citet{zhu2018msmo} incorporated multimodal images and \citet{li2020vmsmo} utilized videos to help better summarization.
	 These works are typically designed for one specific modality of side information,  while a more generally useful summarization framework should be able to process different modalities of information in a flexible way.
    Hence, in this paper, we target to address a general summarization framework that can flexibly unify different modal side information with the input document to generate better summaries.

    There are two main challenges in this task.
    The first challenge comes from the different modalities of side information.
    \textcolor{black}{Regardless of the presented \emph{format} of side information, a summarization model needs to align and unify it with the document into the same semantic space. }
 The other challenge lies in the fact that the side inputs can contain information from various \emph{aspects}, and the model should recognize the aspects useful for summarization. 
    In the first case in Figure~\ref{fig:intro}, only if the summarization model can connect the visual information ``earth'' and ``launching'' to the textual information can it generate the informative summary.
    In the second case in Figure~\ref{fig:intro}, the query describes the question from computer and safety aspects, which should be the focus when making a summary.

    In this work, we propose a Unified-Modal Summarization model with Side information (USS) to tackle the above challenges.
    Firstly, we propose to use topics as the bridge to model the relationship between the main document and the side information.
    Topics are a subject or theme of documents or videos, and traditional works employ topics as cross-document semantic units to bridge different documents \cite{cui2021topic}.
    Moreover, we observe that topics can also be an information bridge for multi-modal inputs.
    For instance, in the first case in Figure~\ref{fig:intro}, we can use topics ``\textit{aerospace}'' and ``\textit{nature}'' to relate the videos with the summary text.
    Hence, in this work, we expand the topic modeling from single-modal to multi-modal for unifying the main document and various types of side information.
    For the second challenge, apart from the limited side-document pairs, we utilize rich non-paired side and document inputs in the collected datasets, and propose a cross-modal contrastive learning module to align the \textcolor{black}{main document and side information} 
    into a unified semantic space.  
    Concretely, in our model, we first introduce a unified topic model (UTM) to learn the latent topics of the target summary by using   the main document and the side information to predict the topic distributions of the summary.
    Since UTM aims to predict the topic distribution of the target summary, it does not rely on the specific modality attributes of the input.
    Based on the learned topics, we construct a graph encoder to model the relationship between the  \textcolor{black}{main document and side} inputs.
    In this topic-aware graph encoder, we let information from two sources flow through different channels, i.e., by direct edges and indirect edges through topics.
	In the decoding process, we propose a hierarchical decoder that attends to multi-granularity nodes in the graph guided by the topics.
    Moreover, the triplet contrastive learning mechanism pushes the paired \textcolor{black}{document and side}
    representations closer and unpaired representations far away from each other, so as to enhance the model's capability of  understanding \textcolor{black}{the main document and side information}. 

	Our contributions can be summarized as follows:
	
	$\bullet$ We propose a general summarization paradigm that can take advantage of different types of side information in a flexible way 
 to enhance summarization performance.
	
	$\bullet$ To model the interaction between various inputs and unify them into the same semantic space, we propose a unified topic model and a triplet contrastive learning mechanism.

	$\bullet$ Empirical results demonstrate that our proposed approach brings substantial improvements over strong baselines on benchmark datasets.

	\section{Related Work}

  \textbf{Summarization with Side Information.}
	Simply relying only on the main body of the document for summarization cues is challenging \cite{zhao2019abstractive,zhu2021leveraging,mao2022muchsum,lima2022extractive}.
	In fact, articles in real-world applications often have side information that is beneficial for summarization.
	A series of works utilized textual side information such as image captions \cite{narayan2017neural}, questions \cite{Koupaee2018WikiHowAL,Deng2019JointLO,deng2020multi}, prototype summaries \cite{Gao2019Abstractive}, citation papers~\cite{chen123capturing,an2021enhancing}, timeline information~\cite{chen2019Timeline}, and prototype templates~\cite{gao2019write}.
	Recently, research on multimodal understanding gets popular, and the benefits of using visual information on summarization have also been explored.
  \citet{gao2021standard} provided a survey on side information-aware summarization.
    Side information-aware summarization can also be regarded as a kind of multi-document summarization.
    \citet{cui2021topic,zhou2021entity} introduced topic and entity information in the summarization process, respectively.
	Different from previous works which either take visual or textual side input, we propose a general framework that can be flexibly applied with different types of side inputs.


	  \textbf{Topic Modeling }
	Neural topic modeling (NTM) was first proposed by \citet{miao2017discovering}, which assumes a Gaussian distribution of the topics in a document.
\citet{liu2019neural,fu2020document,yang2020graph,xie2021graph} further explored it in the summarization task in the text domain.
Specifically, \citet{cui2021topic} employed NTM to jointly discover latent topics that can act as cross-document semantic units to bridge different documents and provide global information to guide the summary generation.
\citet{liu2021topic} proposed topic-aware contrastive learning objectives to implicitly model the topic change and handle information scattering challenges for the dialogue summarization task.
	In this work, we come up with a unified topic model to fit in the \textcolor{black}{unified-modal setting, which requires discovering latent topics beyond single-modal text input. }

	 \textbf{Contrastive Learning.}
    Contrastive learning is used to learn representations by teaching the model which data samples are similar or not.
    Due to its excellent performance on self-supervised and semi-supervised learning, it has been widely used in natural language processing.
    \citet{lee2020contrastive} generated positive and negative examples by adding perturbations to the hidden states.
\citet{cai2020group} augmented contrastive dialogue learning with group-wise dual sampling.
    It has also  been utilized in caption generation~\cite{mao2016generation}, summarization~\cite{gao2019write,liu2021topic,liu2021simcls,bui2021self}, dialog generation ~\cite{he2022unified}, machine translation~\cite{bonab2019simulating,yang2019reducing} and so on.
    In this work, we use contrastive learning to unify multimodal information in the summarization task.

	\section{Model}
	
	In this section, we first define the task of unified summarization with side information, then describe our USS model in detail.
	
	\subsection{Problem Formulation}

Given the main document  $X^d$ and its side information $X^s$,  we assume there is a ground truth summary $Y = (y_1, y_2, \dots, y_{N_y})$.
To be specific, the document $X^d$ is represented as a sequence of words $(x^d_1, x^d_2, \dots, x^d_{N_d})$.
The side information can be in textual or visual formats.
For textual side information, it is represented as $X_s=(x^s_1, x^s_2, \dots, x^s_{N_s})$, and for visual side information, we use $X_s$ to denote the images.
$N_d$ and $N_s$ are the word number or the image number in a document or side information.
Given $X^d$ and $X^s$, our model generates a summary $\hat{Y}= \{\hat{y}_1, \hat{y}_2, \dots, \hat{y}_{N_{\hat{y}}}\}$.
Finally, we use the difference between the generated $Y$ and the gold $\hat{Y}$ as the training signal to optimize the model parameters.
	
	\subsection{Overview}
\begin{figure*}[t]
		\centering
		\includegraphics[width=0.95\linewidth]{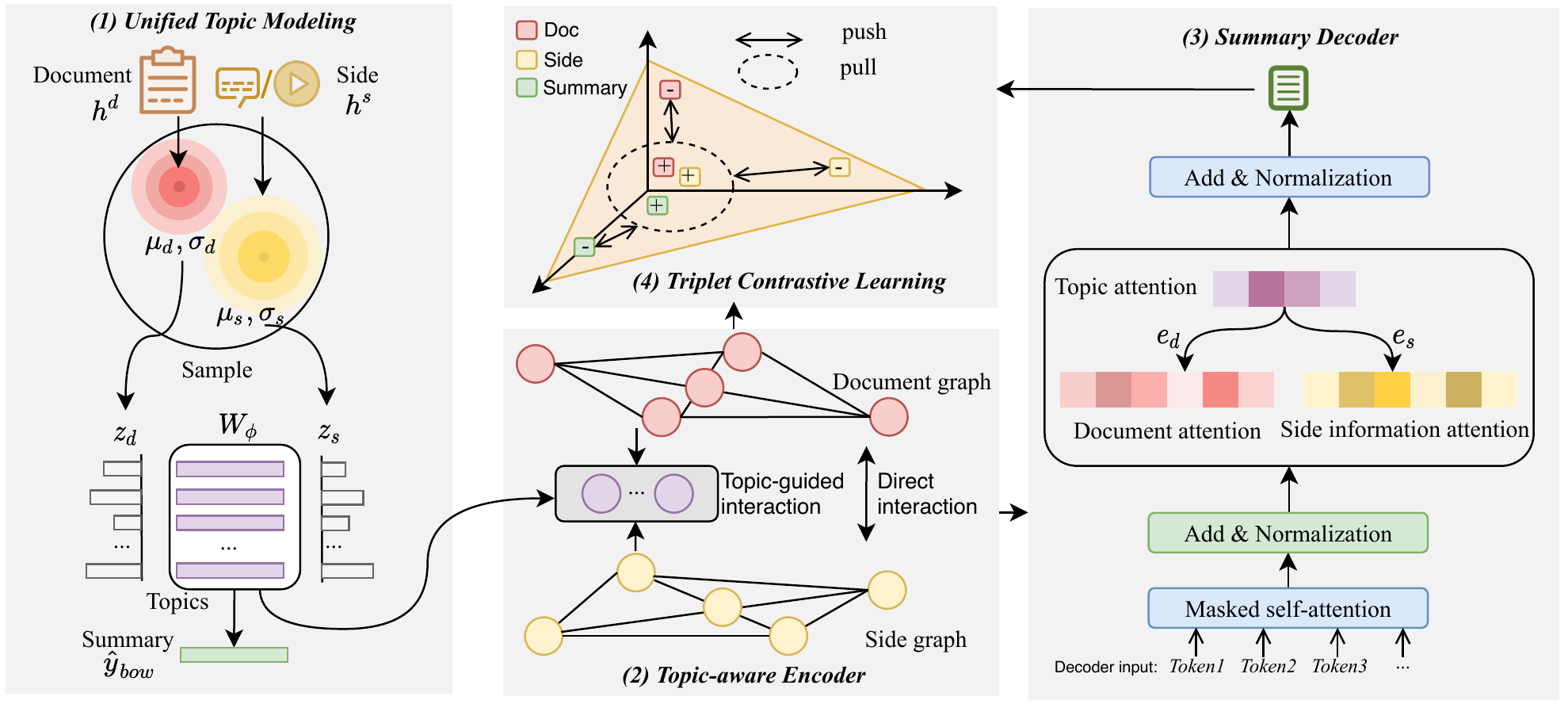}
		\caption{Overview of USS, which consists of four parts: 
		(1) Unified Topic Modeling (left) jointly learns latent topics from both inputs; (2) Topic-aware Graph Encoder (bottom) relates the document to the side information; (3) Summary Decoder (right) with hierarchical topic-aware attention mechanism; and (4) Triplet Contrastive Learning (top) aligns the multiple inputs and outputs into a unified semantic space.}
		\label{fig:model}
	\end{figure*}

	Our model is illustrated in Figure~\ref{fig:model}, which follows the Transformer-based encoder-decoder architecture.
	We augment the encoder with a unified topic modeling network (\S~\ref{subsec:embedding}) which learns the latent topic representations from source inputs and target summary, based  on which a topic-aware graph encoder (\S~\ref{subsec:memory}) builds graphs for the document and side input, and models their relationship through the learned topics.
    Correspondingly, we design a summary decoder (\S~\ref{decoder}) which generates the summary with a topic-aware attention mechanism.
To better represent the representations from different spaces, we also design a triplet contrastive learning module  (\S~\ref{subsec:contras}) to align the paired multimodal information into the same space.
	
	\subsection{Unified Topic Modeling}
	\label{subsec:embedding}
	
	We first use  a unified topic model (UTM) to establish the relationship between the document and side information.
    The model takes inspiration from the neural topic model (NTM) \cite{miao2017discovering} which only applies for textual inputs.
    We first introduce the NTM, and how we adapt NTM to grasp the semantic meanings of multimodal inputs.

    Overall, NTM assumes the existence of $K$ underlying topics throughout the inputs.
    Concretely, NTM encodes the bag-of-word term vector of the input to a topic distribution variable, based on which it reconstructs the bag-of-word representation.
    In the reconstruction process, the topic representations can be extracted from a projection matrix.
    In our UTM, instead of reconstruction, we aim to predict the bag-of-word vector of the target summary based on the two inputs.
    The benefits are threefold.
    Firstly, we no longer require the input to be in textual format and can encode various modal semantic meanings of the inputs into the distribution variable.
    Secondly, we can  preserve the most salient information from the inputs, instead of keeping them all, which is consistent with the information filtering attribute of the summarization task.
    Lastly, the combination of topic modeling on document and side input can better fit the target summary topic distribution.

    Concretely, we first process the document $X^d$ into the bag-of-word representation $h^d \in \mathbb{R}^{|V|}$, where $|V|$ is the vocabulary size.
    The same is true for the side information when it is in the textual format, leading to $h^s$.
    When the side information is images or videos, we use EfficientNet \cite{tan2019efficientnet} to obtain the vector representation, also denoted as $h^s$.
   We then employ an MLP encoder to estimate their exclusive priors $\sigma^*$ and $\mu^*$, which are used to generate the topic variables of the two inputs through a Gaussian softmax:
   \begin{align}
    \sigma^*=f_{\sigma}\left(h^*\right), \mu^* &=f_{\mu}\left(h^*\right), \\
    z^* \sim  \mathcal{N}\left(\sigma^*, \mu^{*2}\right),
    \end{align}
    where $*$ can be $d$ or $s$, $f_{\sigma}(.)$ and $f_{\mu}(.)$ are   neural perceptrons with ReLU activation. $\mathcal{N}(.)$ is a Gaussian distribution.  
    $z^* \in \mathbb{R}^K$ are the latent topic variables of the document or the side information.
    
     Given the topic variables $z^d$ and $z^s$, UTM predicts the bag-of-word representation of target summary, i.e., $y_{bow}$:
     \begin{align}
        \theta=\operatorname{softmax}\left(z^d+z^s\right),  \hat{y}_{bow}=\operatorname{softmax}\left(\theta\mathrm{W}_{\phi}\right).
     \end{align}
     We add the topic variables of the two inputs together to include information from two sources, as well as to emphasize the salient information that is shared between both sides.
     Based on the topic distribution $\theta$ we construct the bag-or-word of target summary $y_{bow}$.
In this process, the weight matrix of $\mathrm{W}_{\phi} \in \mathbb{R}^{K\times |V|}$ can be regarded as the topic-word relationship, where $\mathrm{W}_{\phi}^{i,j}$ indicates the weight of the $j$-th word in the $i$-th topic, and $K$ is the topic number.
$\theta \in \mathbb{R}^{K}$ reflects the proportion of each topic, and higher $\theta_i$ score means the $i$-th topic is more important. 
We will take advantage of this distribution to determine the main topics of each case in the next section.
    
    The objective function is to simultaneously minimize the Wasserstein distance between $p(z^*)$ and $q(z^* \mid h^*_x)$, and maximize the constructing probability of $y_{bow}$:
\begin{align*}
        \mathcal{L}_{UTM}=\textstyle \sum_{* \in \{d,s\}} \operatorname{W}(p(z^*) \| q(z^* \mid h^*_x))-\mathbb{E}_{q(z)}[\log p(y_{bow} 
        \mid z^*)],
    \end{align*}
 where $p(z^*)$ is the standard Gaussian distribution.
    We employ the Wasserstein distance instead of traditional KL-divergence since the former is proved to be superior to the latter by experiments \cite{tolstikhin2017wasserstein}.

	\subsection{Topic-aware Graph Encoder}
	\label{subsec:memory}

	\textbf{Graph Construction.}
	Since we have extracted the salient topic distribution of the two inputs, we can use them as bridges to let two information sources interact with each other.
We thus design a topic-aware graph encoder where we model the relation between document and side inputs through different channels, i.e., by direct edges and indirect edges through topics. 
By direct edges, we let information flow globally in the graph, while by indirect edges, the document communicates specific information with side input under different topics.

    \textbf{Node Initialization.}
    For both inputs, we use the Transformer encoder \cite{vaswani2017attention} or the EfficientNet model to encode each document or image independently to capture the contextual information.
    We first introduce the Transformer architecture in detail, and we will also propose variations of the attention mechanism.
    Generally, Transformer consists of a stack of token-level layers to obtain contextual word representations in the document or side information.
We take the document to illustrate this process.

For the $l$-th Transformer layer, we first use a fully-connected layer to project word state $h^{d,l-1}_{x_{i}}$ into the query, i.e., $Q^{l-1}_{i} = F_q(h^{d,l-1}_{x_{i}})$.
    For self-attention mechanism, the key and value are obtained in a similar way: i.e., $K^{l-1}_{i} = F_k(h^{d,l-1}_{x_{i}})$, $V^{l-1}_{i} = F_v(h^{d,l-1}_{x_{i}})$.
	Then,  the updated representation of $Q$ is formed by linearly combining the entries of $V$ with the weights:
		\begin{align}
	\alpha_{i,j} =\frac{\exp \left(Q^{l-1}_{i} K^{l-1}_{j}\right)}{\sum_{n=1}^{N_d} \exp \left(Q^{l-1}_{i} K^{l-1}_{n}\right)}, 
	a_{i} =\sum_{j=1}^{N_d} \frac{\alpha_{i,j} V^{l-1}_{j}}{\sqrt {d_e}}, 
	\end{align}
	where $d_e$ stands for hidden dimension.
	The above process is summaized as $\text{MHAtt}(h^{d,l-1}_{x_{i}},h^{d,l-1}_{x_{*}})$, where $*$ denotes index from $1$ to $N_d$.
	Then, a residual layer takes the output of self-attention sub-layer as the input:
\begin{align}
\hat{h}_{x_{i}}^{d,l-1}&=\text{LN}\left(h_{x_{i}}^{d,l-1}+\text{MHAtt}\left(h_{x_{i}}^{d,l-1}, h_{x_{*}}^{d,l-1}\right)\right),\label{transformer1} \\
h_{x_{i}}^{d,l}&=\text {LN }\left(\hat{h}_{x_{i}}^{d,l-1}+\operatorname{FFN}\left(\hat{h}_{x_{i}}^{d,l-1}\right)\right),\label{transformer2}
\end{align}
where FFN is a feed-forward network with an activation function, LN is a layer normalization \cite{ba2016layer}.

    The final output representations for words in the document and textual side information are denoted as $\{h^d_{1},...,h^d_{N_d}\}$ and $\{h^s_{1},...,h^s_{N_s}\}$.
    For images in visual side information, we also denote them by  $\{h^s_{1},...,h^s_{N_s}\}$.
    As for the topic nodes, we use the intermediate parameters $W_{\phi}$ learned from UTM as raw features to build topic representations $H^o = f_{\phi}(W_{\phi})$, where the $i$-th row of $H^o \in \mathbb{R}^{K \times d_e}$, denoted as $h^o_i$, is a topic vector with predefined dimension $d_e$.
    $f_{\phi}(.)$ is a neural perceptron with ReLU activation.

    \textbf{Graph Encoding.}
    The document and side graphs communicate with each other through \textit{topic-guided} and \textit{direct} interactions.
    The topic-guided interaction starts from the learning of document representations and the side representations, and then the topic representations.
    The direct interaction only  updates the document and side nodes.
    We omit the layer index here for brevity.
    
\begin{figure}
    \centering
    \includegraphics[width=1\columnwidth]{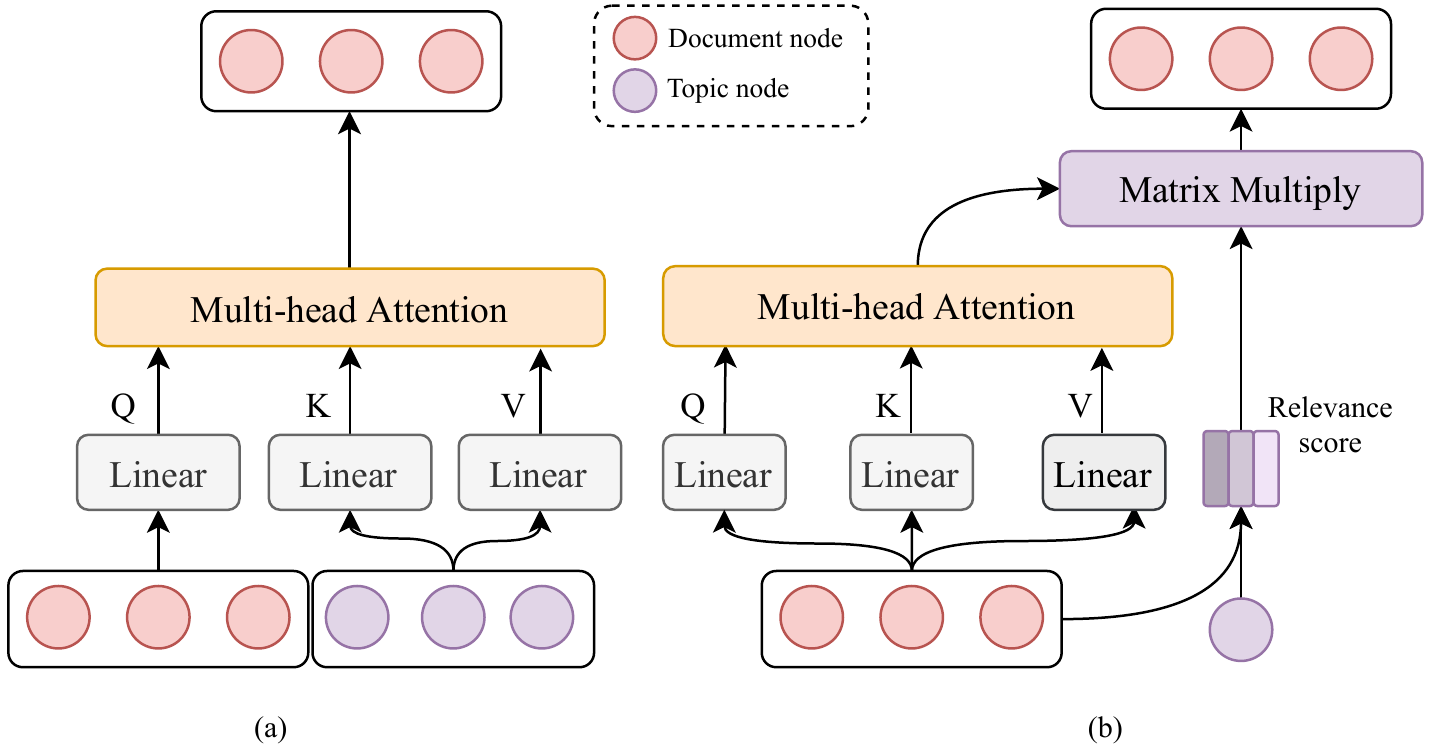}
    \caption{
          (a) Cross attention mechanism for document and topic nodes.
          (b) Topic-guided attention mechanism, which captures semantic information across the document and side information under the guidance of the topics.
    }
    \label{fig:attn}
\end{figure}

    Concretely, in the topic-guided interaction, the \textit{document} and \textit{side information} representations are updated from three sources. 
Take the document nodes for example, they are updated by 
(1) performing self-attention across document nodes;
(2) performing cross-attention to obtain the topic-aware document representations, as shown in Figure~\ref{fig:attn}(a); and
(3) performing our designed topic-guided attention mechanism, as shown in Figure~\ref{fig:attn}(b). 
This mechanism starts with the application of self-attention mechanism on the document nodes: $\dot{h}^d_i=\text{MHAtt}\left(h^d_{i}, h^d_{*}\right)$.
Then, taking the topic representation $h^{o}=\textstyle \sum^{K}_{i=1}h^o_i$ as condition, the attention score $ \beta_{i}$ on each original document representation $h^d_{i}$ is calculated as:
\begin{align}
    \beta_{i}^d &=\text{softmax}\left(\text{FFN}\left(h^d_i(h^{o})^T\right)\right).
\end{align}
The topic-aware document representation is $\dot{h}^d_i$ weighted by $\beta_{i}^d$, denoted as $\beta_{i} \dot{h}^d_i$.
In this way, we highlight the salient part of the two inputs under the guidance of the topics.
Last, a feed-forward network is employed to integrate three information sources.

The \textit{topic} representation is updated by performing (1) self-attention and (2) cross-attention on the adjacent document and side nodes. 
In the cross-attention, the topic representation is taken as the query, and the document and side representations are taken as the keys and values.
Lastly, a feed-forward network integrates two information sources to obtain the updated topic representation.

    Aside from communicating the graphs through topics, we also have a direct interaction that concatenates all document and side nodes in the graph and then apply a self-attention mechanism.

The topic-aware and direct interactions are processed iteratively, and we denote the final updated representations for document, side information, and topics as $\hat{h}^d \in \mathbb{R}^{N_d \times d_e}$, $\hat{h}^s \in \mathbb{R}^{N_s \times d_e}$, and $\hat{h}^o \in \mathbb{R}^{K \times d_e}$.

    \subsection{Summary Decoder}
    \label{decoder}

    Since the decoder needs to incorporate the information from multiple sources in the graph encoder, we design a hierarchical decoder that firstly focuses on the topics and then attends to inputs. 
    This topic-guided mechanism indicates which topics should be discussed in each decoding step.
    Our hierarchical decoder follows the style of Transformer,
    and we omit the layer index next for brevity.
    
    For each layer, at the $t$-th decoding step, we first apply the masked self-attention on the  summary embeddings (MSAttn), obtaining the decoder state $\tilde{g}_t$.
    The masking mechanism ensures that the prediction of the position $t$ depends only on the known output of the position before $t$:
    \begin{align*}
        \tilde{g}_t=\text {LN }\left(g_t+\text{MSAttn}\left(g_t, g_*\right)\right).
    \end{align*}
    Based on $ \tilde{g}_t$ we compute the cross-attention scores over topics:
    \begin{align}
    z_{o,t}=\text{ReLU}([ \tilde{g}_t W_a(\hat{h}^{o} W_b)^T]),
    \end{align}
    where $W_a, W_b \in \mathbb{R}^{d_e \times d_e}$, $z_{o,t} \in \mathbb{R}^K$.
    We then use the topic attention to guide the attention on the other two graphs, where the topics can be regarded as an indicator of saliency.
    Taking the main document for example, we incorporate $z_{o,t}$ with similarity weight $e_d$ to obtain the document attention weight $z_{d,t} \in \mathbb{R}^{N_d}$:
    \begin{align}
        e_d&=\text{FFN} (\hat{h}^{o}(\hat{h}^{d})^T),\\
        z_{d,t}&=\textstyle z_{o,t}  e_d,
    \end{align}
    where $e_d \in \mathbb{R}^{K \times N_d}$ is the similarity matrix between the topics and document. 
    In a similar way, we obtain the attention weights $z_{s,t} \in \mathbb{R}^{N_s}$ on the side information.
    
    The attention weights $z_{o,t}$, $z_{d,t}$, and $z_{s,t}$ are then used to obtain the context vectors $c_{o,t}$, $c_{d,t}$, and $c_{s,t}$, respectively.
    Take the topics as an example:
    \begin{align}
        c_{o,t}= z_{o,t} \hat{h}^{o}.
    \end{align}

    These context vectors, treated as salient contents summarized from various sources, are concatenated with the decoder hidden state $g_t$ to produce the distribution over the target vocabulary:
    \begin{align}
        P_t^{\text {vocab }}=\text{Softmax}\left(W_{d}\left[g_{t} ; c_{o,t};c_{d,t};c_{s,t}\right]\right).
    \end{align}
    All the learnable parameters are updated by optimizing the negative log likelihood objective function of predicting the target words:
    \begin{align}
        \mathcal{L}_{s}=-\textstyle \sum_{t=1}^{N_{y}} \log P_t^{\text {vocab}}\left(\hat{y}_{t}\right).
    \end{align}

\begin{table*}[t!]
    \caption{ Comparison with other baselines when side information is in text.
    		All our ROUGE scores have a 95\% confidence interval of at most $\pm$0.28  as reported by the official ROUGE script.
    		Numbers in \textbf{bold} mean that the improvement to the best baseline is statistically significant (a two-tailed paired t-test with p-value \textless 0.01).
      `-'  indicates unavailability.}
    	\begin{center}
    	\resizebox{0.8\textwidth}{!}{	\begin{tabular}{l|cccc|cccc|cccc}
    			\bottomrule
    			\multirow{3}*{}
    			&& CNN & & & & WikiHow &&&& VMSMO \\
    			&  R1 &  R2 &  RL& BS&  R1 &  R2 &  RL &BS& R1 &  R2 &  RL& BS\\
    			\hline
    			Lead3 & 29.1 & 11.1 &25.9  &85.4& 24.4 & 5.5 &15.6 & 84.7 & 16.2 & 5.3 & 13.9 & 84.5\\
    			
    			BERTSumEXT& 31.2 & 12.2 & 27.8 & 86.3 &30.4 & 8.6 & 28.3 & 85.7 & 27.3 & 9.6 & 23.4& 85.7\\
       SideNet& 30.7 & 11.7 & 27.2& 86.2 & 28.4 & 6.2 & 25.9 & 85.4 &- &- & -& -\\
    			\hline
    		
    			BERTSumABS & 31.3 & 11.6 & 28.7  & 86.5 &35.9 & 13.9 & 34.8& 86.9 & 27.8 & 10.1 & 24.7 & 85.9 \\		BERTSumABS-\textit{concat} & 31.7 & 11.9 & 29.1 & 86.6 & 35.7 & 14.0 & 34.2 & 86.7 &- & - & - & -\\
    			SAGCopy & 31.9 & 11.8 & 28.9& 86.5 & 36.1 & 13.7 & 33.9 & 86.6& -&-& -  \\ 
    			EMSum & 31.9 & 12.6 & 29.7 & 86.8 & 35.4 & 13.8 & 34.6  &  86.7 &- & -& -& - \\
    			TG-MultiSum & 32.2 & 12.7 & 29.9 & 86.8 &36.2 & 14.6 & 35.2& 87.0 &- &-& - & -\\
    			\hline
           OFA & 31.6 & 11.4  &  28.4 & 86.1 & 34.6 & 13.2 & 33.8 & 86.3  & 30.5  & 13.5 & 28.4 & 86.9 \\
    			MOF& - & - & -  && - & - & - && 30.8 & 13.4 & 28.4 & 86.9\\
    			VMSMO& - & - & -  && - & - & - && 31.2 & 13.8 & 28.5 & 87.1\\

    			\hline
    			USS & \textbf{33.9} & \textbf{14.2} & \textbf{31.3} & \textbf{87.1} & \textbf{37.7} & \textbf{15.8} & \textbf{36.5} & \textbf{87.3} & \textbf{32.7} & \textbf{15.1} & \textbf{30.0}& \textbf{87.9}  \\
    			USS w/o unified topic modeling & 32.5& 13.0 &30.3 &86.8& 36.5 &15.0 & 35.4& 87.0 & 30.9 & 12.8 & 27.9 & 87.4   \\
    			USS w/o graph encoder & 32.3 & 12.9 & 30.0& 86.7 &36.2 & 14.8 & 35.5& 87.1 & 29.5 & 11.6 & 26.5 & 87.3 \\
    			USS w/o contrastive learning &  32.8 & 13.2& 30.6& 86.9 & 36.7 & 15.3& 35.8 & 87.1 &30.4 & 13.1 & 27.8 & 87.5  \\
    			\bottomrule
    		\end{tabular}}
    	
    	\end{center}
    	
    	\label{tab:compare-baseline}
    \end{table*}
  
	\subsection{Triplet Contrastive Learning}
	\label{subsec:contras}
	The challenge of unifying different modalities is to align and unify their representations at different levels.
	In this section, we propose a triple contrastive learning mechanism that determines whether the textual and visual representations match each other.
 We can utilize the large-scale non-paired text corpus and image collections to learn more generalizable textual and visual representations, and improve the capability of vision and language understanding.

	As shown in the fourth part in Figure~\ref{fig:model}, the main idea is to let the representations of the paired images or text close to each other in the semantic space while the non-paired be far away.
	For the positive sample construction, we apply mean pooling on the representations of $\hat{h}^{d}_{*}$ as the overall representation $D$ for the document, and $S$ for side information in the same way.
	The final decoder state of generator $g_{N_{\hat{y}}}$ is taken as the overall representation $G$ for the generated summary, as it stores all the accumulated information.
	For the negative sample construction, we randomly sample a negative side input, document, or the generated summary from the same training batch for each case.
	Note that different from the positive pairs, the sampled side and texts are encoded individually without graph encoders as they mainly carry weak correlations.
	In this way, we can create  positive examples $\mathcal{X}^{d+}$ consisting of paired document-side samples $(D,S)$, $\mathcal{X}^{s+}$ consisting of paired side-generation $(S,G)$, and $\mathcal{X}^{g+}$ consisting of paired document-generation $(G,D)$.
	Negative examples are denoted as $\mathcal{X}^{d-}$, $\mathcal{X}^{s-}$, and $\mathcal{X}^{g-}$, respectively.
	
	Based on these positive and negative pairs, the following contrastive loss $\mathcal{L}_{triplet}$ is utilized to learn detailed semantic alignments across vision and language:
	\begin{align*}
	    \mathcal{L}_{triplet}=\mathbb{E}_{A,B}\left[-\log \frac{\sum_{\left(A^{+}, B^{+}\right) \in \mathcal{X}^{\{g+, d+, s+\}}} \exp \left(d\left(A^{+}, B^{+}\right) / \tau\right)}{\sum_{\left(A^{*}, B^{*}\right) \in \mathcal{X}^{\{g^*,d^*,s^*\}}} \exp \left(d\left(A^{*}, B^{*}\right) / \tau\right)}\right],
	\end{align*}
	where $\tau$ denotes the temperature parameter, $A,B \in \{D,S,G\}$, and $d(A,B)$ denotes cosine similarity between the two representations. 

\section{Experiments}
    \subsection{Dataset}
    We evaluated our model on three public summarization datasets with side information:
    (1) CNN dataset is collected by \citet{narayan2017neural} from CNN webpage, which associates the document-summary pairs with the image captions.
    The dataset contains 90,266/1,220/1,093 training/validation/test samples.
    Note that this dataset is different from the CNN/DM dataset~\cite{hermann2015teaching}.
    (2) WikiHow dataset \cite{Koupaee2018WikiHowAL} is an abstractive summarization dataset collected from a community-based QA website, in which each sample consists of a query, a document as the answer to the question, and a short summary of the document.
    We use the provided script and obtain more than 170,000 pairs, with 168128/6000/6000 training/validation/test samples.
    (3) VMSMO dataset \cite{li2020vmsmo} is a multimodal summarization dataset with videos as side information. 
    The average video duration is one minute and the frame rate of the video is 25 fps. 
    Overall, there are 184,920 samples in the dataset, which is split into a training set of 180,000 samples, a validation set of 2,460 samples, and a test set of 2,460 samples.
    The average document length, side information length, and summary length of CNN and WikiHow datasets are 763, 71, 50 words, 579, 10, 62 words, respectively.
    The document and summary length for VMSMO is 97 and 11 words.
     We chose these datasets to examine the flexibility of our model on different  formats of side information.
    
    \subsection{Baselines}
    Our extractive baselines include: 
    
    \textbf{Lead3} produces the three leading sentences of the document as the summary as a baseline.
    
    \textbf{SideNet} \cite{narayan2017neural} consists of an attention-based extractor with attention over side information. 
    
     \textbf{BERTSumEXT} \cite{liu2019text} is  an extractive summarization model with pretrained BERT encoder that is able to express the semantics of a document and obtain representations for its sentences. 
     It only takes the document as input.
     
    Abstractive single-document and multi-document summarization baseline models include:
      
    \textbf{BERTSumABS} \cite{liu2019text} is an abstractive summarization system built on BERT$_\text{base}$ with a new designed fine-tuning schedule.
    It only takes the document as input.
    We also have a \textbf{BERTSumABS-\textit{concat}} that concatenates the textual side information with the original document.

   \textbf{SAGCopy}~\cite{xu2020self} is an augmented Transformer with a self-attention guided copy mechanism.
   
    \textbf{EMSum} \cite{zhou2021entity} is an entity-aware model for abstractive multi-document summarization with BERT encoder.
    
    \textbf{TG-MultiSum} \cite{cui2021topic} is a multi-document summarizer with topics act as cross-document semantic units.

    The above two multi-document summarization baselines take the textual side input as the second document. 
    We also compare our model with multimodal summarization baselines including:
    
    \textbf{MOF} \cite{zhu2020multimodal} is a summarization model with a multimodal objective function with the guidance of multimodal reference to use the loss from the summary generation and the image selection. 
    
    \textbf{VMSMO} \cite{li2020vmsmo} is a dual interaction-based multimodal summarizer with  multiple inputs.
    The four above models are all equipped with BERT$_\text{base}$ encoder for fairness.

 \textbf{OFA} \cite{wang2022ofa} is a recent unified paradigm for multimodal pretraining.
 We adapt it for the side-aware summarization setting, where we directly concatenate the document and side representations encoded by OFA.
 We choose $\text{OFA}_{\text{base}}$ version for fairness.

    \subsection{Evaluation Metrics}
For both datasets, we evaluated by standard ROUGE-1 (R1), ROUGE-2 (R2), and ROUGE-L (RL)~\cite{lin2004rouge} on full-length F1, which refer to the matches of unigram, bigrams, and the longest common subsequence, respectively.
We then used BERTScore (BS) \cite{zhang2019bertscore} to calculate a similarity score between the summaries based on their BERT embeddings.

\citet{schluter2017limits} noted that only using the ROUGE metric to evaluate generation quality can be misleading. 
Therefore, we also evaluated our model by human evaluation.
Concretely, we asked three PhD students  proficient in  English to rate 100 randomly sampled cases generated by models from the CNN and WikiHow datasets which cover different domains.
The setting follows \cite{liu2022end} with four times larger evaluation scale.
The evaluated baselines are EMSum, TG-MultiSum, and OFA which achieve top performances in automatic evaluations.

Our first evaluation quantified the degree to which the models can retain the key information following a question-answering paradigm \cite{Liu2019HierarchicalTF}.
	We created a set of questions based on the gold-related work and examined whether participants were able to answer these questions by reading generated text. 
	The principle for writing a question is that the information to be answered is about factual description, and is necessary for the summarization. 
	Two annotators wrote three questions independently for each sampled case. 
	Then they together selected the common questions as the final questions that they both consider to be important. 
	Finally, we obtained 147 questions, where correct answers are marked with 1 and 0 otherwise. 
	Our second evaluation study assessed the overall quality of the related works by asking participants to score them by taking into account the following criteria: \textit{Informativeness} (does the related work convey important facts about the topic in question?), \textit{Coherence} (is the related work coherent and grammatical?), and \textit{Succinctness} (does the related work avoid repetition?).
	The rating score ranges from 1 to 3, with 3 being the best. 
	Both evaluations were conducted by another three PhD students independently, and a model’s score is the average of all scores.

    \subsection{Implementation Details}
    All models were trained for 200,000 steps on NVIDIA A100 GPU.
We implemented our model in Pytorch and OpenNMT \cite{klein2017opennmt}.
    For neural-based baselines except OFA and our model, we used the `bert-base' or `bert-base-chinese' versions of BERT for fair comparison. 
    Both source and target texts were tokenized with BERT's subwords tokenizer.
    Our Transformer decoder has 768 hidden units and the hidden size for all feed-forward layers is 2,048.
    In all abstractive models, we applied dropout with probability 0.1 before all linear layers; label smoothing \cite{szegedy2016rethinking} with smoothing factor 0.1 was also used. 
    For CNN dataset, the encoding step is set to 750 for a document and 70 for side information.
    The minimum decoding step is 30, and the maximum step is 50.
    For WikiHow dataset, the four parameters are set to 600, 10, 30, 65.
    For the Chinese VMSMO dataset, the parameters are 200, 125, 10, 50, where 125 is the encoded frame number.
     The video frames are selected every 25 frames to ensure the continuity of the images, similar to \cite{li2020vmsmo}.
    We used Adam optimizer as our optimizing algorithm.
    We also applied gradient clipping with a range of $[-2,2]$ during training.
    During decoding, we used beam search size 5, and tuned the $\alpha$ for the length penalty \cite{wu2016google} between 0.6 and 1 on the validation set; we decoded until an end-of-sequence token is emitted and repeated trigrams are blocked.
    Our decoder applies neither a copy nor a coverage mechanism, since we also rarely observe issues with out-of-vocabulary words in the output; moreover, trigram-blocking produces diverse summaries managing to reduce repetitions.
    On VMSMO, the video frames are selected every 25 frames to ensure the continuity of the images, similar to \cite{li2020vmsmo}.
    We selected the 5 best checkpoints based on performance on the validation set and report averaged results on the test set.

    \subsection{Experimental Results}
    \label{overall}

    \textbf{Automatic Evaluation.}
    The performance comparison is shown in Table~\ref{tab:compare-baseline}.
    Firstly, we can see that the attributes of the three datasets vary. 
    CNN is a news dataset with a pyramid structure, where Lead3 and extractive methods achieve higher performance than other datasets.
    Secondly, combining side information by simple concatenation cannot make full use of it, as we can see that the performance of BERTSumABS-\textit{concat} does not improve significantly compared with BERTSumABS.
    Incorporating side information by multi-document summarization structures is a better way utilize side information, but they cannot be applied in the multimodal scenario, and their improvements are also limited.
    Thirdly, the recent multimodal pretrained baseline OFA achieves relatively good performance on VMSMO, but has borderline performance on single-modal datasets CNN and WikiHow.
    This is consistent with the previous observation that OFA has better performance on cross-modal tasks \cite{wang2022ofa}.
    Specifically, OFA has trouble when generating long text, which leads to a performance drop when the target is relatively long.
    Finally, our USS model obtains consistently better performance on all three datasets.
    Specifically, USS achieves 2.0/1.6/1.7/0.3 improvements on R1, R2, RL, and BERTScore compared with one of the latest baseline EMSum on the CNN dataset, and obtains 1.4/1.0/1.3/0.7 improvements on the VMSMO dataset compared with  TG-MultiSum.

    \begin{table}[t]
    \centering
    \small
        \caption{QA performance, Informativeness (Info), Coherence (Coh), and Succinctness (Succ) comparison by human evaluation.}
    \begin{tabular}{@{}lcccc@{}}
    \toprule
    & QA(\%) & Info & Coh & Succ\\
    \midrule
    OFA  & 34.6 & 2.14 & 2.01 & 2.03 \\
    EMSum &  37.4 & 2.39 & 2.13 & 2.26 \\
    TG-MultiSum &  38.1 & 2.42 & 2.15 & 2.28 \\
    USS & \textbf{42.8 }& \textbf{2.54} & \textbf{2.27} & \textbf{2.41} \\
    \bottomrule
    \end{tabular}

    \label{tab:human_evaluation}
    \end{table}

    \textbf{Human Evaluation.}
    As shown in Table~\ref{tab:human_evaluation}, on both evaluations, participants overwhelmingly prefered our model.
    The kappa statistics are 0.42, 0.49, and 0.45 for Info, Coh, and Succ respectively, indicating the moderate agreement between annotators. 
    All pairwise comparisons among systems are statistically significant using a two-tailed paired t-test for strong significance for $\alpha$ = 0.01.
    We also provide examples of the system output in Table~\ref{tab:case_study}.
    We can see that with the side information showing the figure of the main character, the lottery result, and the mobile phone, USS successfully captures the gist information that ``a man post a lottery ticket on social media'' in the generated summary.
    For BERTSumABS-\textit{concat} and VMSMO, they miss key information such as ``where he post the lottery'' and ``how quickly the lottery was falsely claimed''.

    \begin{CJK*}{UTF8}{gkai}
    \begin{table*}[t]
        \centering
        \small
         \caption{Examples of the generated summary by baselines and USS on CNN and VMSMO datasets.
        Unfaithful and redundant information is highlighted in \fph{blue}.
      In the second case, keywords with the same semantics are highlighted in \error{red} and \reph{green}.
        }
        
         \begin{tabular}{ll}
          \toprule
         \multicolumn{2}{p{16cm}}{    \emph{\textbf{Article:}}
         Recently, a citizen of Nantong, Jiangsu,  won a lottery ticket.
         He took photos of the entire lottery ticket and uploaded them to Moments. 
         Unexpectedly, someone else falsely claimed the lottery winnings as his own based on the information on the lottery. 
         The lottery was redeemed within only 35 seconds after the start of the redemption day as investigated by the Sports Lottery Center.
            近日，江苏南通，市民张先生彩票中了奖后，将整张彩票拍照上传了朋友圈，不料被人根据彩票上的信息冒领了奖金。经体彩中心调查当天开始兑奖后仅35秒奖金就被兑走。
          } \\ \hline 
          \multicolumn{2}{p{16cm}}{
            \emph{\textbf{Reference summary:}}
            Too excited to win the lottery, post the lottery in \reph{Moments} and got falsely claimed \error{immediately}
            中奖太兴奋，\reph{朋友圈}晒彩票\error{瞬间}被冒领
            } \\ 
            \hline
              \multicolumn{2}{p{16cm}}{
            \emph{\textbf{OFA:}}  \error{35 seconds} after winning, the lottery was falsely claimed 中奖\error{35秒后}被冒领彩票
         }\\ \hline
                   \multicolumn{2}{p{16cm}}{
            \emph{\textbf{MOF:}}
           Man showed the winning lottery and was falsely claimed in \error{35 seconds} 男子晒中奖\error{35秒}被冒领}\\ \hline

          \multicolumn{2}{p{16cm}}{
          \emph{\textbf{VMSMO:}} Post lottery in \reph{Moments} and get falsely claimed \reph{朋友圈}晒中奖被冒领
           }\\ \hline
              
           \multicolumn{2}{p{16cm}}{
            \emph{\textbf{USS:}}
            Friends from \reph{Moments} falsely claimed the lottery, only \error{35 seconds} after the redemption started
            \reph{朋友圈}冒领彩票，中奖\error{35秒}就被兑走
            }\\ \hline

             \emph{\textbf{Highest three topics:}}\\
             Topic1: old friend 老朋友, Liang family 梁家, WeChat 微信, phone calls 通电话, Brothers 兄弟俩\\
            Topic2: covet 贪图, steal 偷盗, kidnap 拐骗, holocaust 大屠杀, steal everything 抢光\\
            Topic3: prize 奖金, tens of thousands 好几万,  giants 豪门, net flow 净流入, more than 100 million yuan 亿余元
            \\
            
            \hline
            \textit{\textbf{Side information (sampled images from video)}}:\\
            \multicolumn{2}{p{16cm}}{
            \begin{minipage}{0.15\textwidth}
               \includegraphics[scale=0.25]{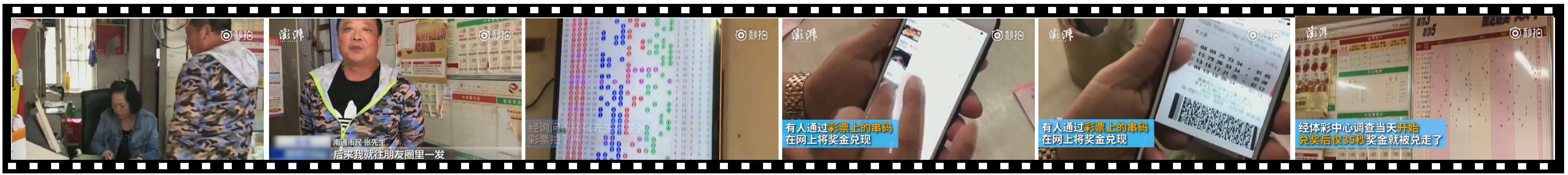}
               
            \end{minipage}} \\
          \bottomrule
        \end{tabular}

        \label{tab:case_study}
    \end{table*}
    \end{CJK*}

    \section{ANALYSIS AND DISCUSSION}
    
    \subsection{Ablation Study}
    We conducted ablation tests to assess the importance of the topic modeling, graph encoder, and triplet contrastive learning.
    For USS w/o unified topic modeling, only the traditional neural topic model (NTM) is applied to the textual document to obtain the topic representations.
    For USS w/o graph encoder, there are no topic-related interactions, and the outputs from the topic modeling are directly used for decoding.
    The ROUGE score results are shown in the last block of Table~\ref{tab:compare-baseline}.
    All ablation models perform worse than USS in terms of all metrics, which demonstrates the preeminence of USS.
    Concretely, graph encoder makes a great contribution to the model, improving the performance on CNN by 1.3 in terms of the R2 score, and improving the R2 score by 1.0 on the WikiHow dataset.
    Contrastive learning also contributes to the model, bringing 0.7 RL improvements on the CNN dataset.

    \begin{figure}%
        \centering
        \subfigure{{\includegraphics[width=3.8cm]{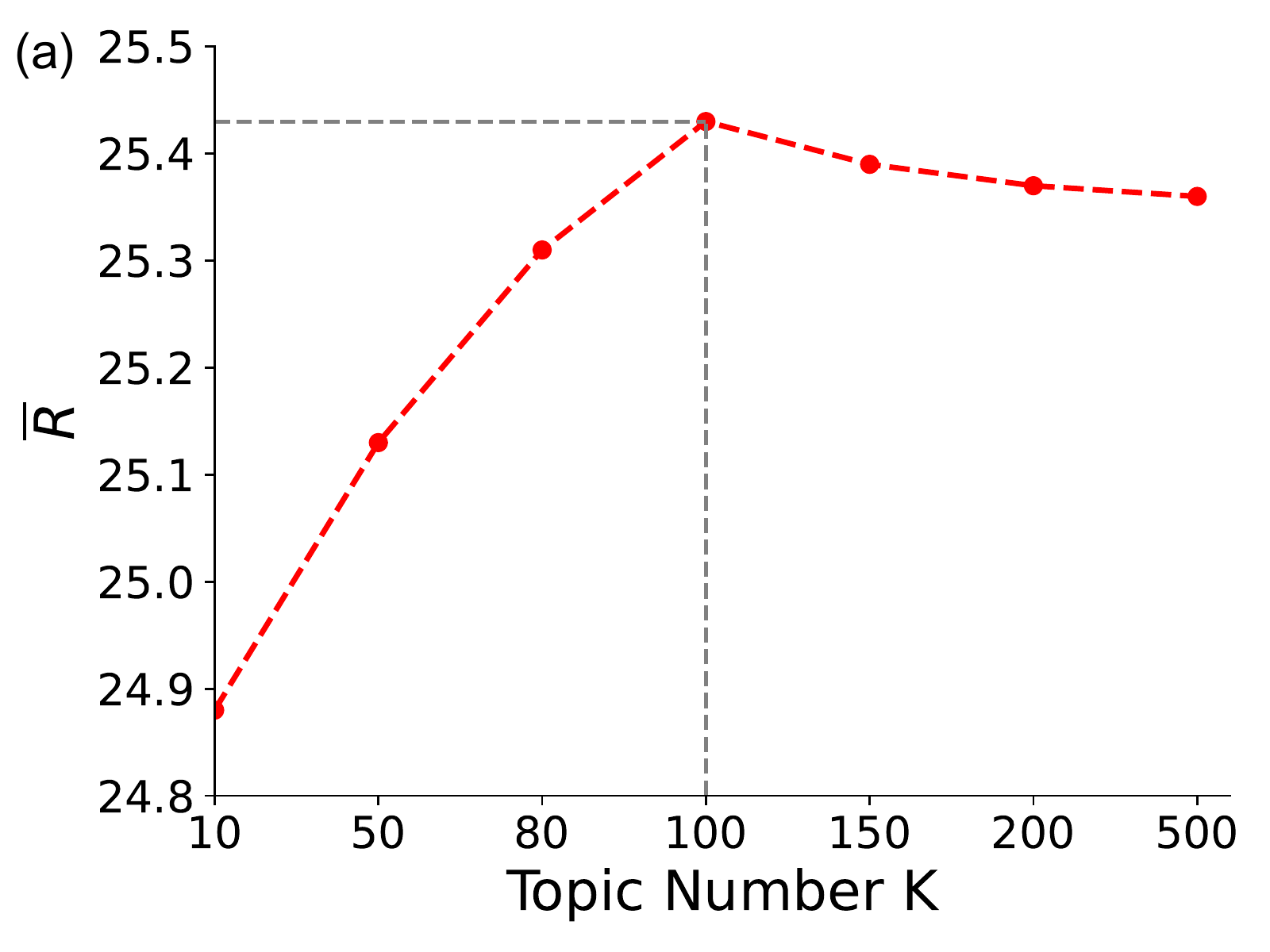} }}%
        \subfigure{{\includegraphics[width=3.8cm]{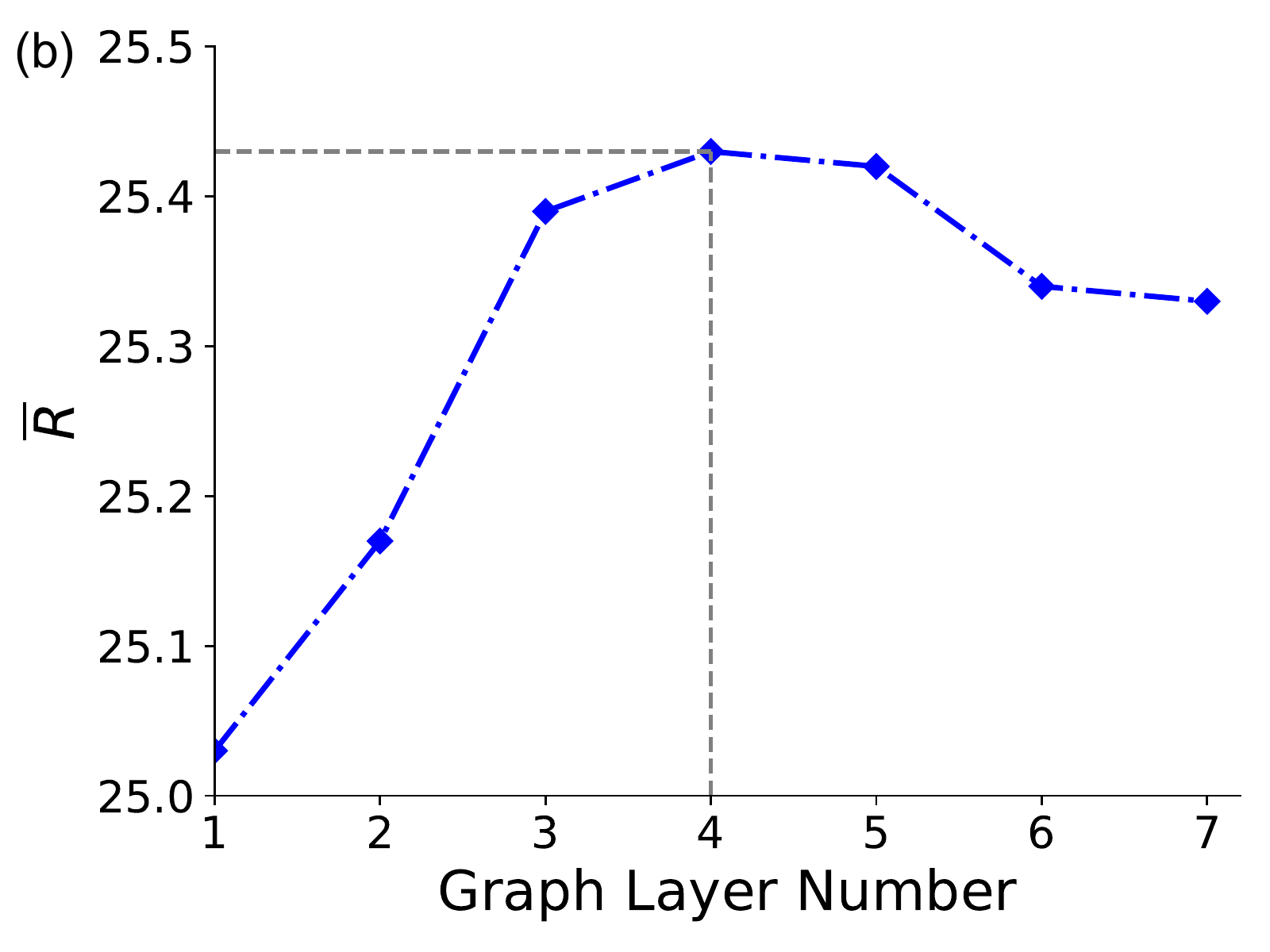} }}%
        \caption{(a) Relationships between the number of topics and $\overline{R}$ (the average of RG-1, RG-2 and RG-L) . 
	Best viewed in color. (b) Relationships between the number of graph layer and $\overline{R}$. }%
        
        \label{hyper}%
    \end{figure}

    We further conduct experiments on VMSMO to probe into the impact of two important parameters, i.e., the topic number $K$ and the graph layer number $L$. 
From Figure \ref{hyper}, we can see that for both experiments, the ROUGE scores increase with the topic and layer number, to begin with. 
After reaching the upper limit it begins to drop.
Note that with only one graph layer our model outperforms the best baseline, which demonstrates that our topic-aware graph module is effective. 
Hence, we set the default topic number to 100 and the graph layer number to 4.

    \begin{table}
    \small
    	\caption{Coherence score $C_v$ and inferred topic words of different topic models.
	\textcolor{blue}{Blue text} denotes repetition or non topic words.}
    \begin{tabular}{c|c|c}
      \toprule
      Models& $C_v$ &  Sampled words\\
      \hline
      GSM & \makecell[c]{} 0.392 & \makecell[l]{political, growth, economy, europe,\\ \textcolor{blue}{according}, world, states, \textcolor{blue}{better},\\ opportunity} \\
      \hline
      W-LDA & \makecell[l]{} 0.462 & \makecell[l]{mcconnell, consensus, electorate, \\reduction, repeal, partisan, \textcolor{blue}{economy},\\ \textcolor{blue}{economies}, growth} \\
      \hline
      USS & 0.493 & \makecell[l]{consumers, billion, growth, economy, \\global, companies, cost, oil, \\ infrastructure, sector} \\
      \bottomrule
    \end{tabular}

    \label{tab:topics}
    \end{table}

 \begin{figure}%
        \centering
        \subfigure{{\includegraphics[width=3.8cm]{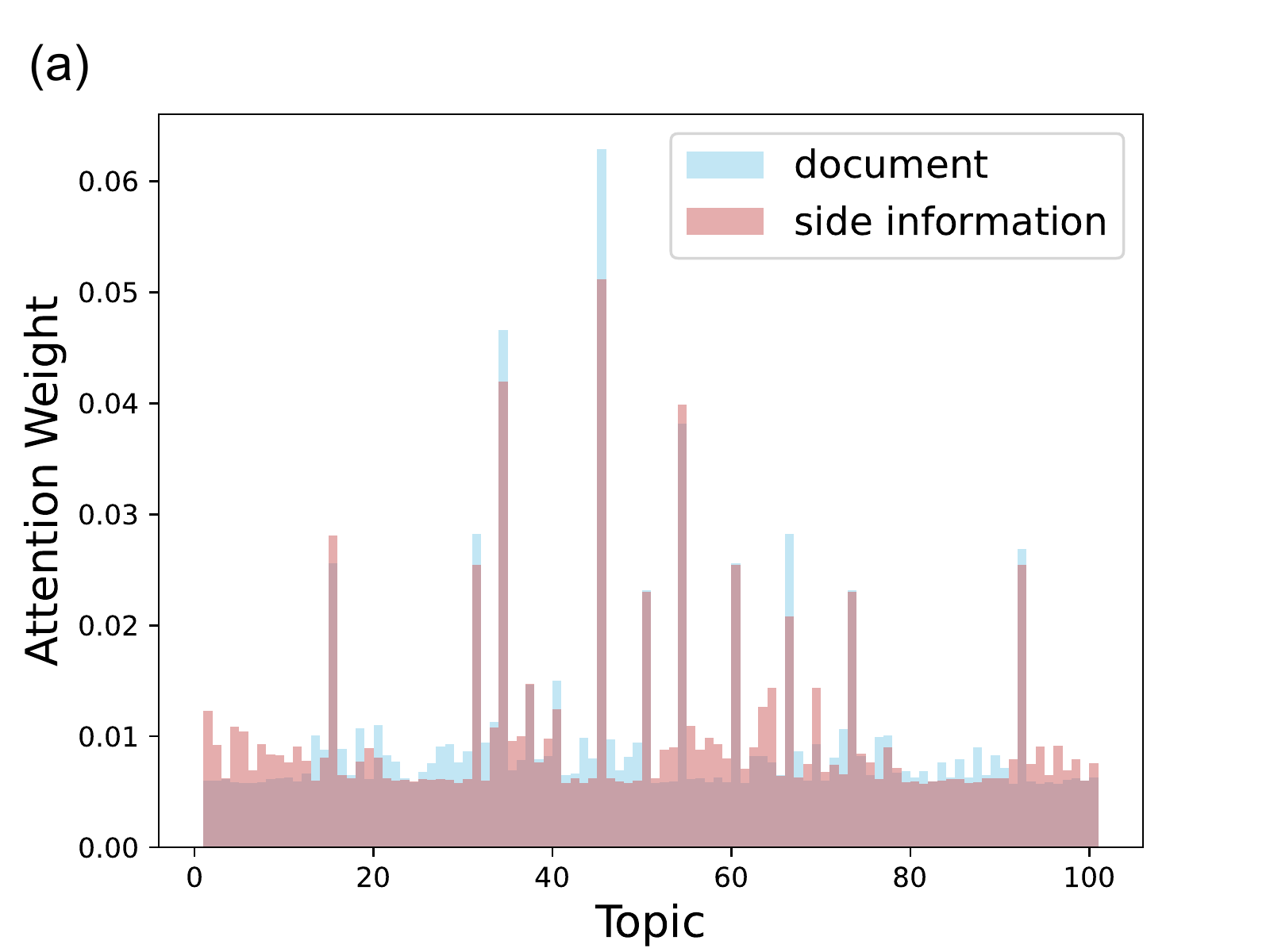} }}%
        \subfigure{{\includegraphics[width=3.8cm]{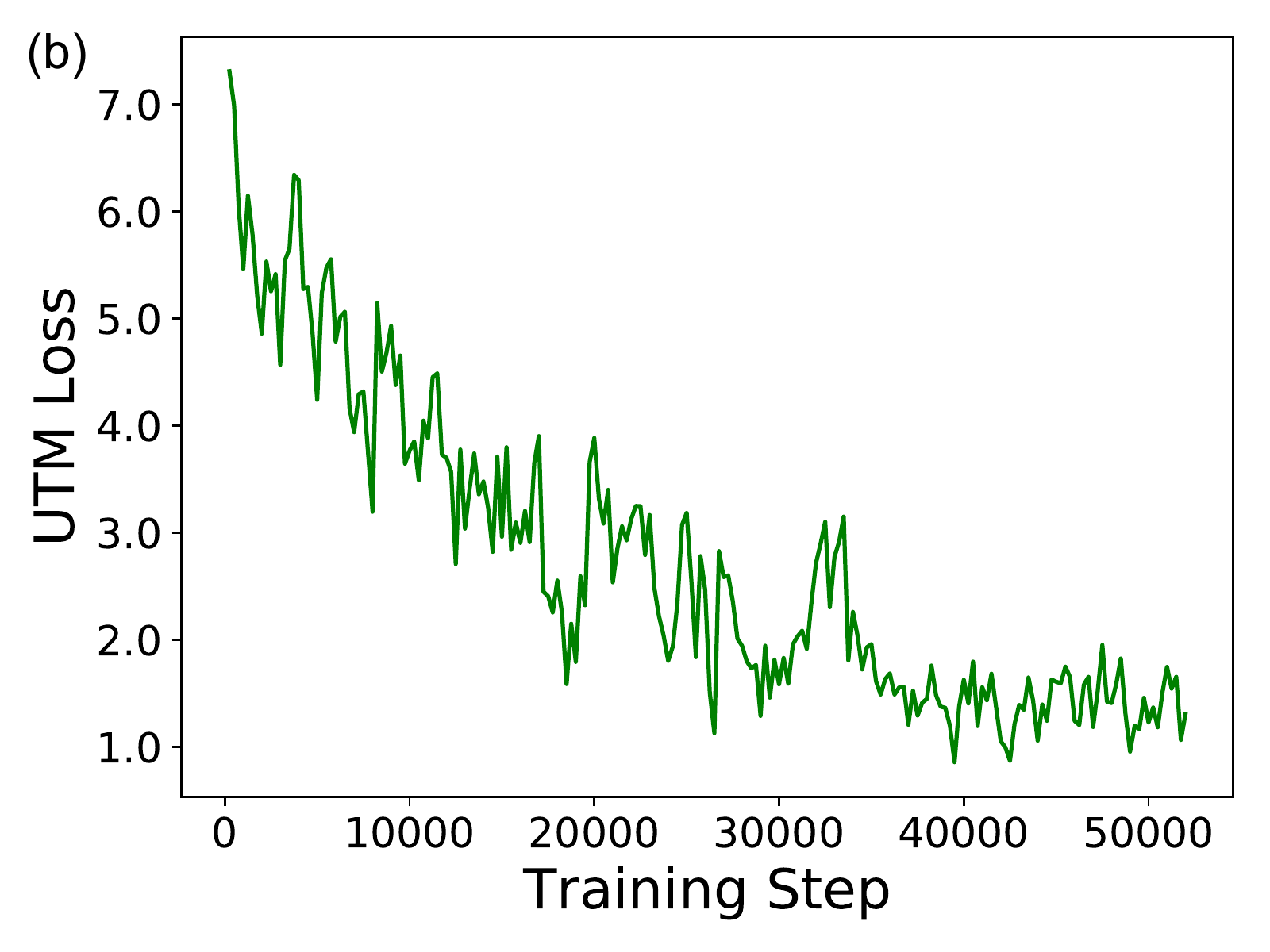} }}%
        \caption{(a) The multi topic distribution of the document and side information. (b) UTM loss ($\mathcal{L}_{UTM}$) curve in the training process. }%
        
        \label{fig:my_label}%
    \end{figure}

       \subsection{Topic Quality Analysis}
    In this subsection, we qualitatively and quantitatively investigate the quality of the selected topics.
    We compared the learned topics from our model with baseline topic models trained on the CNN dataset including (1) GSM \cite{miao2017discovering}, a classic NTM model with VAE and Gaussian softmax, and (2) W-LDA \cite{nan2019topic}, a novel neural topic model in the Wasserstein autoencoders framework.

    In Table~\ref{tab:topics}, we use the coherence score $C_v$ \cite{roder2015exploring} to quantitatively evaluate inferred topics, which has been proved highly consistent with human evaluation. 
    We also show the inferred words for the topic ``economy''.
    It can be seen that our USS outperforms other baselines in terms of the coherence score, and the inferred topic words are more accurate and concentrated. 
    The possible reasons are twofold.
    Firstly, our model incorporates the main and side inputs to predict the topic distribution of the target summary.
    The multiple descriptions of the same content bring more topic clues, and the prediction task that requires reasoning and filtering abilities makes the topic model strong and robust.
    Secondly, the assistant summarization task can boost the performance of topic modeling.
    
     \begin{figure}%
        \centering
        \subfigure{{\includegraphics[width=3.8cm]{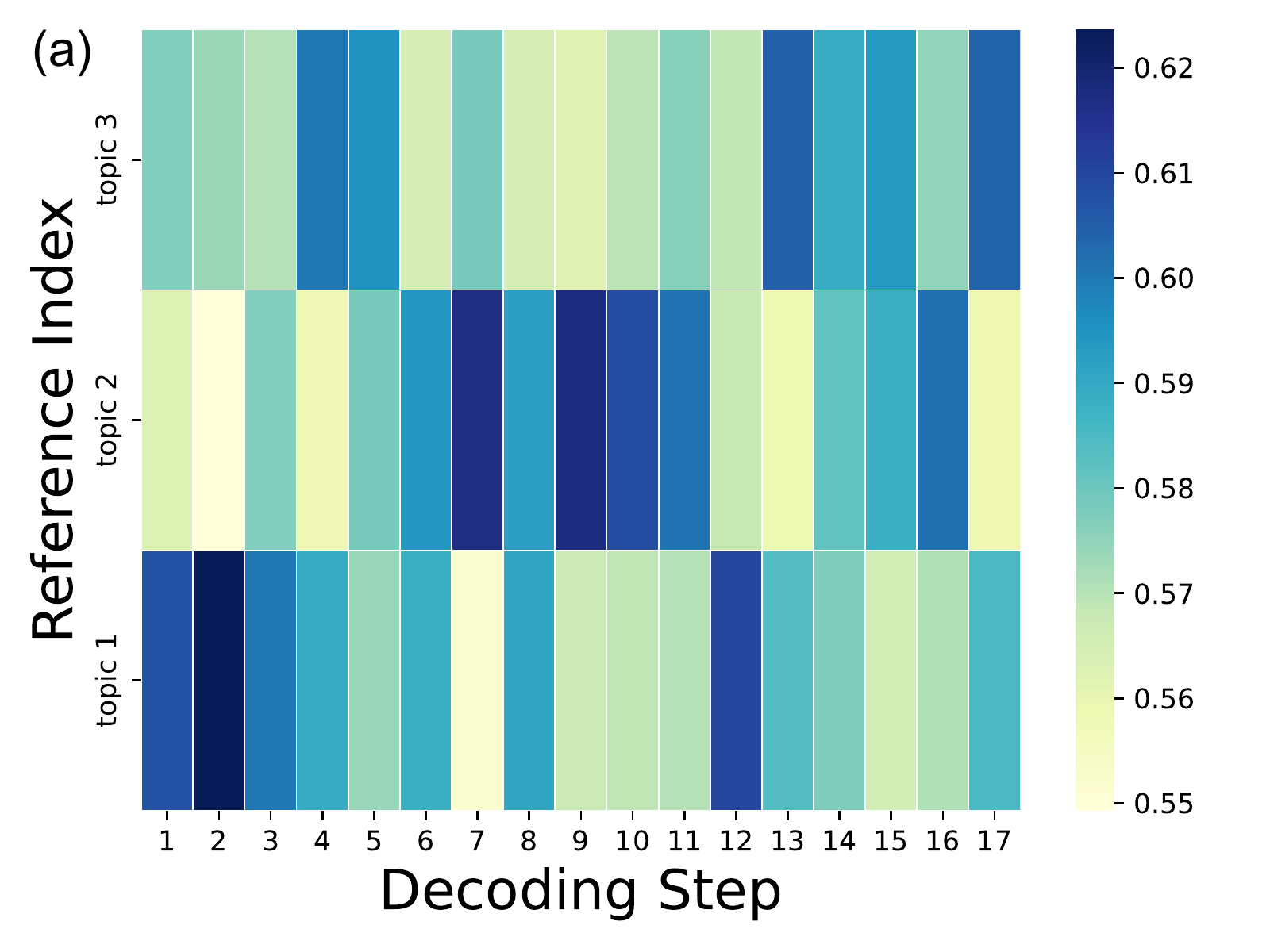} }}%
        \subfigure{{\includegraphics[width=3.8cm]{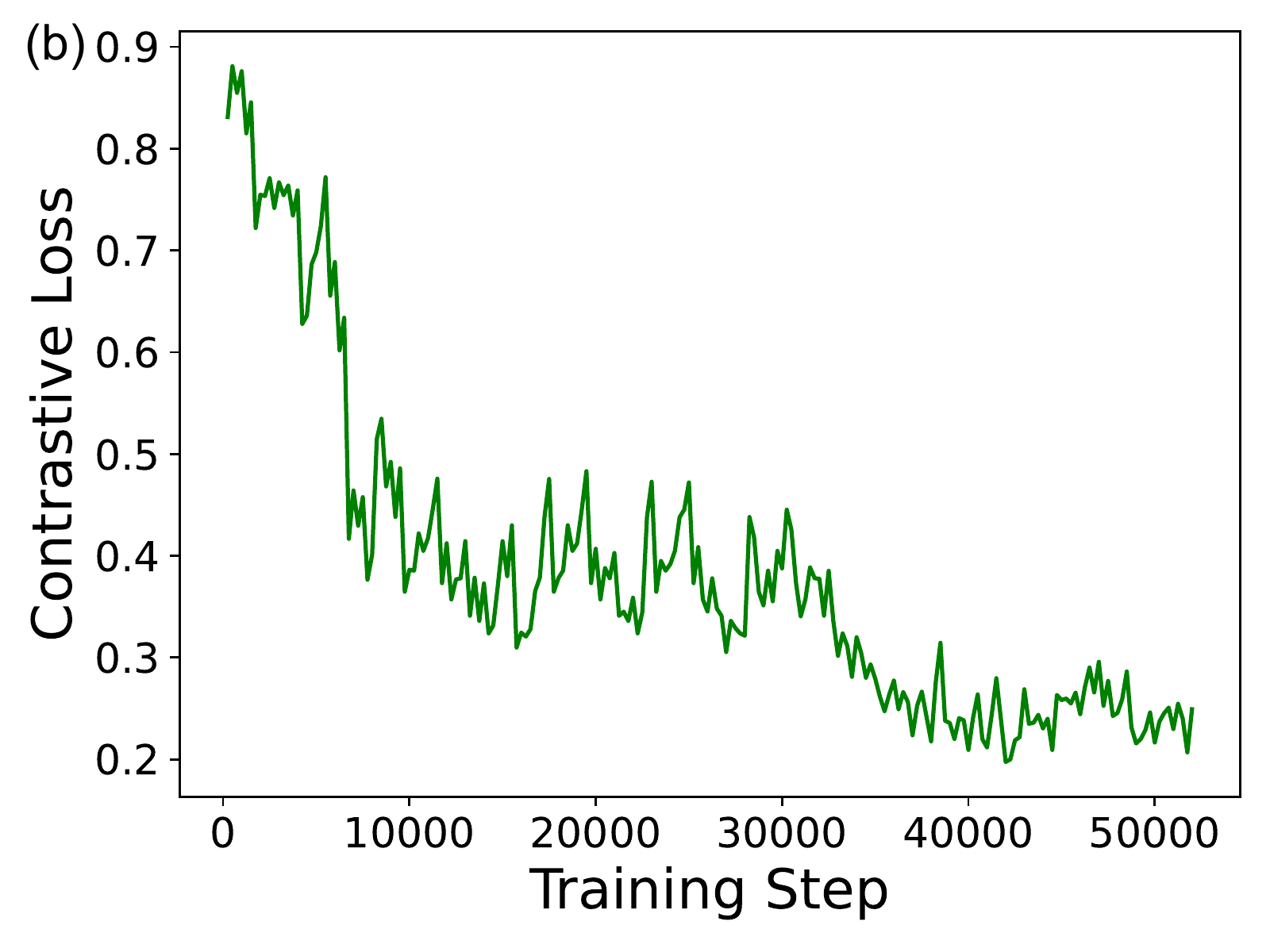} }}%
        \caption{(a) Visualizations of the attention weights on topics. (b) Contrastive learning loss curve ($\mathcal{L}_{triplet}$) in the triplet contrastive learning module. }%
        
        \label{f2}%
    \end{figure}

   \subsection{Effect of Unified Topic Modeling}
   Since we have verified the quality of the topics, we are interested to see the effect of the learned topics on summarization, i.e., how the unified topic modeling helps summarization?

  We first examine from the \textit{encoder} side, where we show the learned topic distributions from two inputs for the case in Table~\ref{tab:case_study} in Figure~\ref{fig:my_label}(a).
   It can be seen that though the document and side information has different topic distributions, generally, they focus on the same important topics, which are related to the ground summary by human evaluation.
   From the statistic view, we draw the loss of $\mathcal{L}_{UTM}$ in Figure~\ref{fig:my_label}(b).
   The curve has a steady downtrend, to begin with, and finally reaches convergence.
   The above observations demonstrate that the topic modeling can grasp the gist of the target summary and the effectiveness of topic modeling.
   
We next examine the topic effectiveness in the summarization process from the \textit{decoder} side.
We  visualize the attention weights $z_{o,t}$ on topics in Figure~\ref{f2}(a) for the same case.
    It can be seen that the topic attention first emphasizes topic 1, and then on topic 2 as well as topic 3. 
    The three topics are shown in Table \ref{tab:case_study}, which are related to ``social media'', ``crime'', and ``finance'', respectively.
    This is consistent with the generated sentence, where the keyword starts from ``Moments'', and then changes to ``falsely claimed redemption''.
    In this way, we can see that the topics play a guidance role when generating summaries.


    \subsection{Contrastive Learning Analysis}
    \label{visual}
    
    We lastly examine the performance of the triplet contrastive learning module by visualizing the contrastive loss curve in Figure~\ref{f2}(b) on VMSMO.
    It can be seen that the loss score fluctuates at the beginning of the training and gradually reaches convergence.
    This phenomenon demonstrates that the generated text, the document, and the side information belonging to the same case are getting closer in the semantic space.
    On the other hand, the unpaired triplets are becoming more distant.
	
\section{Conclusion and limitation}
    In this paper, we proposed a general summarization framework, which can flexibly incorporate various modalities of side information.
	We first proposed a unified topic model to learn latent topic distributions from various modal inputs.
	We then employed a topic-aware graph encoder that relates one input to another by topics.
	Experiments on three public benchmark datasets show that our model produces fluent and informative summaries, outperforming strong systems by a wide margin.


\section*{Acknowledgments}
We would like to thank the anonymous reviewers for their constructive comments. 
The work was supported by King Abdullah University of Science and Technology (KAUST) through grant awards FCC/1/1976-44-01, FCC/1/1976-45-01, REI/1/5234-01-01, and RGC/3/4816-01-01.

\bibliographystyle{ACM-Reference-Format}
\bibliography{sample-base}

\end{document}